\documentclass[11pt,a4paper]{article}
\usepackage[hyperref]{naaclhlt2019}
\usepackage{times}
\usepackage{latexsym}
\usepackage{url}
\usepackage{xfrac}
\usepackage{multirow}
\usepackage{amsmath}
\usepackage{amssymb}
\usepackage{rotating}
\usepackage[colorinlistoftodos]{todonotes}
\newcommand{\myurl}[1]{\href{http://#1}{\tt \nolinkurl{#1}}}

\usepackage{color}
\definecolor{red}{RGB}{255,110,71}
\definecolor{blue}{RGB}{110,130,255}
\definecolor{purple}{RGB}{110,130,255}

\expandafter\def\expandafter\normalsize\expandafter{%
    \normalsize
    \setlength\abovedisplayskip{2pt}
    \setlength\belowdisplayskip{2pt}
    \setlength\abovedisplayshortskip{2pt}
    \setlength\belowdisplayshortskip{2pt}
}

\global \hyphenpenalty 2000
\global \exhyphenpenalty 1000

\newcommand{\svdpmi}{\textsc{SVD$_{\textsc{PPMI}}$}}
\newcommand{\svdwpmi}{\textsc{SVD$_{\textsc{wPPMI}}$}}
\newcommand{\glove}{\textsc{GloVe}}
\newcommand{\sgns}{\textsc{SGNS}}

\aclfinalcopy
 
\author{Johannes Hellrich \hfill Bernd Kampe \hfill Udo Hahn\vspace{8pt}\\
\{\texttt{firstname.lastname}\}\texttt{@uni-jena.de} \vspace{4pt} \\
Jena University Language \& Information Engineering (JULIE) Lab\\
Friedrich-Schiller-Universit{\"a}t Jena, Jena, Germany\\
\myurl{julielab.de}
}

 \definecolor{red}{rgb}{0,0,0}
 
\title{The Influence of Down-Sampling Strategies\\ on SVD Word Embedding Stability}

\begin{document}
\maketitle
\begin{abstract}
The stability of word embedding algorithms, i.e., the consistency of the word representations they reveal when trained repeatedly on the same data set, has recently raised concerns. We here compare word embedding algorithms on three corpora of different sizes, and evaluate both their stability and accuracy. We find strong evidence that down-sampling strategies (used as part of their training procedures) are particularly influential for the stability of \svdpmi -type embeddings. This finding seems to explain diverging reports on their stability and lead us to a simple modification which provides superior stability as well as accuracy on par with skip-gram embeddings.
\end{abstract}

\section{Introduction}
Word embedding algorithms implement the latest form of distributional semantics originating from the seminal work of \citet{Harris54} or \citet{Rubenstein65}. They generate dense vector space representations for words based on co-occurrences within a context window. 
They sample word-context pairs, i.e., typically two co-occurring tokens, from a corpus and use these to generate vector representations of words and their context. 
Changes to the algorithm's sampling mechanism can lead to new capabilities, e.g., processing dependency information instead of linear co-occurrences  \citep{Levy14acl}, or increased performance, e.g., using word association values instead of raw co-occurrence counts \citep{Bullinaria07}. 

Word embedding algorithms commonly down-sample contexts to lessen the impact of high-frequency words (termed `subsampling' in \citet{Levy15}) or increase the relative importance of words closer to the center of a context window (called `dynamic context window' in \citet{Levy15}). The effect of using such down-sampling strategies on accuracy in word similarity and analogy tasks was explored in several papers (e.g., \citet{Levy15}). 

However, down-sampling and details of its implementation also have major effects on the stability of word embeddings (also known as `reliability'), i.e., the degree to which models trained independently on the same data agree on the structure of the resulting embedding space. This problem has lately raised severe concerns in the word embedding community (e.g., \citet{Hellrich16coling,Antoniak18,Wendlandt18}) and is also of interest to the wider machine learning community due to the influence of probabilistic---and thus unstable---methods on experimental results \citep{Reimers17,Henderson18}, as well as replicability and reproducibility \citep[pp.\,63:3--4]{Ivie18}. 

Stability is critical for studies examining the underlying semantic space as a more advanced form of corpus linguistics, e.g., tracking lexical change \citep{Kim14,Kulkarni15,Hellrich18coling}. Unstable word embeddings can lead to serious problems in such applications, as interpretations will depend on the luck of the draw. This might also affect high-stake fields like medical informatics where patients could be harmed as a consequence of misleading results \citep{Coiera18}.


In the light of these concerns, we here evaluate down-sampling strategies by modi\-fying the \svdpmi\ (Singular Value De\-com\-po\-sition of a Positive Pointwise Mutual Information matrix; \citet{Levy15}) algorithm and comparing its results with those of two other embedding algorithms, namely, \glove\ \citep{Pennington14} and \sgns\  \citep{Mikolov13iclr,Mikolov13nips}. Our analysis is based on three corpora of different sizes and investigates  effects on both accuracy and stability. The inclusion of accuracy measurements and the larger size of our training corpora exceed prior work.
We show how the choice of down-sampling strategies, a seemingly minor detail, leads to major differences in the characterization of  \svdpmi\  in recent studies \citep{Hellrich17dh,Antoniak18}. We also present \svdwpmi , a simple modification of \svdpmi\ that replaces probabilistic down-sampling with weighting. What, at first sight, appears to be a  small change leads, nevertheless, to an unrivaled combination of stability and accuracy, making it particularly well-suited for the above-mentioned corpus linguistic applications.
\section{Computational Methodology}

\subsection{Measuring Stability}%
\label{sec:reliability}
Measuring word embedding stability can be linked to older research comparing distributional thesauri \citep{Salton71chapter} by the most similar words they contain for \textcolor{red}{particular} anchor words \citep{Weeds04,Padro14}.
Most stability experiments focused on repeatedly training the \textit{same} algorithm on one corpus \citep{Hellrich16latech,Hellrich16coling,Hellrich17dh,Antoniak18,Pierrejean18taln,Chugh18}, whereas \citet{Wendlandt18} quantified stability by comparing word similarity for models trained with \textit{different} algorithms. We follow the former approach, since we deem it more relevant for ensuring that study results can be \textcolor{red}{replicated or reproduced}.

\textit{Stability} can be quantified by calculating the overlap between sets of words considered most similar in relation to pre-selected anchor words. Reasonable metrical choices are, e.g., the Jaccard coefficient \citep{Jaccard12} between these sets \citep{Antoniak18,Chugh18}, or a percentage based coefficient \citep{Hellrich16latech,Hellrich16coling,Wendlandt18,Pierrejean18taln}.
We here use $j@n$, i.e., the Jaccard coefficient for the $n$ most similar words. It depends on a set $M$ of word embedding models, $m$, for which the $n$ most similar words (by cosine) from a set $A$ of anchor words, $a$, as provided by the 'most similar words' function $\mathrm{msw}(a,n,m)$, are compared:
\vspace*{5pt} 
\begin{equation} \label{eq:rel:jaccard}
\begin{gathered}
j@n := \dfrac{1}{|A|}\sum_{a \in A}  \frac{ | \bigcap_{m \in M} \mathrm{msw}(a,n,m) | }{ | \bigcup_{m \in M} \mathrm{msw}(a,n,m) | }
\end{gathered}
\end{equation} 
\subsection{\svdpmi\ Word Embeddings}
The \svdpmi\ algorithm from \citet{Levy15}   generates word embeddings in a three-step process. First, a corpus is transformed to a word-context matrix listing co-occurrence frequencies. Next, the frequency-based word-context matrix is transformed into a word-context matrix that contains word association values. Finally, singular value decomposition (SVD; \citet{Berry92,Saad03}) is applied to the latter matrix to reduce its dimensionality and generate word embeddings.

Each token from the corpus is successively processed in the first step by recording co-occurrences with other tokens within a symmetric window of a certain size. For example, in a token sequence $\dots , w_{i-2}, w_{i-1},  w_{i},  w_{i+1},  w_{i+2}, \dots $, with $w_i$ as the currently modeled token, a window of size 1 would be concerned with $w_{i-1}$ and  $w_{i+1}$ only. Down-sampling as described by \citet{Levy15} increases accuracy by ignoring certain co-occurrences while populating the word-context matrix (further details are described below). A word-context matrix is also used in \glove , whereas \sgns\ directly operates on sampled co-occurrences in a streaming manner.

Positive pointwise mutual information (PPMI) is a variant of pointwise mutual information \citep{Fano61,Church90}, independently developed by \citet{Niwa94} and \citet{Bullinaria07}. PPMI measures the ratio between observed co-occurrences (normalized and treated as a joint probability) and the expected co-occurrences (based on normalized frequencies treated as individual probabilities) for two words $i$ and $j$ while ignoring all cases in which the observed co-occurrences are fewer than the expected ones:
\vspace*{5pt} 
\begin{equation} 
PPMI(i,j) := 
\begin{cases}
0 & \text{if  } \frac{P(i,j)}{P(i)P(j)} < 1 \\
log(\frac{P(i,j)}{P(i)P(j)}) & \text{otherwise}
\end{cases}\label{method:eq:ppmi}
\end{equation}

Truncated SVD reduces the dimensionality of the vector space described by the PPMI word-context matrix $M$. SVD  factorizes $M$ in three special\footnote{
$U$ and $V$ are orthogonal 
matrices containing so called singular vectors. 
$\Sigma$ is a diagonal 
matrix containing singular values.
} matrices, so that $M = U \Sigma V^{T}\!$. Entries of $\Sigma$ are ordered by their size, allowing to infer the relative importance of vectors in $U$ and $V$. This can be used to discard all but the highest $d$ values and corresponding vectors during truncated SVD, so that $M_d = U_d \Sigma_d V^{T}_d \approx M$. Both \glove\ and \sgns\ start with randomly initialized vectors of the desired dimensionality $d$ and have thus no comparable step in their processing pipeline. However, \citet{Levy14nips} showed \sgns\ to perform as an approximation of SVD applied to a PPMI matrix.
\subsection{Down-sampling}
\label{sec:downsampling}
Down-sampling by some factor requires both a formal expression to define the factor, as well as a strategy to perform down-sampling according to this factor---data can either be sampled probabilistically or weighted (see below). The following set of formulae is shared by \sgns\ and \svdpmi , whereas \glove\ uses a distinct one. 

Distance-based down-sampling depends on the distance between the currently modeled token $w_i$ and a second token $w_j$ in a token sequence (such as the above example). The distance $d$ between $w_i$ and  $w_j$ is given as:
\vspace*{5pt} 
\begin{equation}\label{eq:distance}
d(w_i, w_j) := |j - i|
\end{equation}


\vspace*{5pt} 
To increase the effect of the nearest---and thus \mbox{assumedly} most salient---tokens both \svdpmi\  and \sgns\ down-sample words based on this distance with a distance factor, $df$ ($s$ being the size of the window used for sampling):
\vspace*{5pt} 
\begin{equation}\label{eq:sgns:window}
df(w_i, w_j) := \frac{s + 1 - d(w_i, w_j)}{s}
\end{equation}

\vspace{5pt}
To limit the effect of high-frequency words---likely to be function words---both algorithms also down-sample words according to a frequency factor 
($f\!\!f$), which compares each token's relative frequency $r(w)$ with a threshold $t$: 
\vspace*{5pt} 
\begin{equation}\label{eq:sgns:singlefreq}
f\!\!f(w) := \begin{cases} 
\sqrt{t/r(w)} & \text{if }  r(w) > t \\
1 & \text{otherwise}
\end{cases}
\end{equation}

\vspace{5pt}
The frequency down-sampling factor for the co-occurrence of two tokens $w_i$ and $w_j$ is then given by the product of their down-sampling factors, i.e., the probabilities are treated as being independent:
\vspace*{-10pt} 
\begin{equation}\label{eq:sgns:freq}
f\!\!f(w_i,w_j) :=f\!\!f(w_i) \cdot f\!\!f(w_j) 
\end{equation}

\vspace{5pt}
The strategy used to apply these down-sampling factors can affect accuracy and, especially, stability, as can the decision not to apply them at all. These down-sampling processes can either be probabilistic, i.e., each word-context pair is processed with a probability given by $df(w_i, w_j) \cdot f\!\!f(w_i,w_j)$, or operate by weighting, i.e., for each observed co-occurrence only a fraction of a count according to the product of $df$ and $f\!\!f$ is added to the word-context matrix. \sgns\ uses probabilistic down-sampling, \glove\ uses weighting and  \svdpmi\ by \citet{Levy15} allows for probabilistic down-sampling or no down-sampling at all. As SVD itself is non-probabilistic\footnote{
Assuming that a non-stochastic SVD algorithm \citep{Halko11} is used, as in \citet{Levy15}.
} \citep[chs.\,6.3\,\&\,7.1]{Saad03} any instability observed for \svdpmi\ must be caused by its probabilistic down-sampling. We thus suggest \svdwpmi , i.e., SVD of a PPMI matrix with weighted entries, a  simple modification which uses fractional counts according to $df(w_i, w_j) \cdot f\!\!f(w_i,w_j)$.  
As shown in Section \ref{sec:results}, this modification is beneficial for both accuracy and stability. 
\section{Corpora}
The corpora used in most stability studies are relatively small. For instance, the largest corpus in \citet{Antoniak18} contains 15M tokens, whereas the corpus used by \citet{Hellrich17dh} and the largest corpus from \citet{Wendlandt18} each contain about 60M tokens. \citet{Pierrejean18taln} used three corpora of about 100M words each. Two exceptions are \citeauthor{Hellrich16coling} (2016a,b)  using relatively large Google Books Ngram corpus subsets \citep{Michel11} with 135M to 4.7G n-grams,
 as well as \citet{Chugh18} who investigated the influence of embedding dimensionality on stability based on three corpora with only 1.2--2.6M tokens.\footnote{
Size information from personal communication.}

We used three different English corpora as training material: the 2000s decade of the Corpus of Historical American English (COHA; \newcite{Davies12}), the English News Crawl Corpus (NEWS) collected for the 2018 WMT Shared Task\footnote{\scalebox{.98}[1.0]{
\myurl{statmt.org/wmt18/translation-task.html}}} and a Wikipedia corpus (WIKI).\footnote{
To ease replication, we used a pre-compiled 2014 Wikipedia corpus: \myurl{linguatools.org/tools/corpora/wikipedia-monolingual-corpora/}}   COHA contains 14k texts and 28M tokens, NEWS 27M texts and 550M tokens, and WIKI 4.5M texts and 1.7G tokens, respectively. COHA was selected as it is commonly used in corpus linguistic studies, whereas NEWS and WIKI serve to gauge the performance of all algorithms in general applications. The latter two corpora are far larger than common in stability studies, making our study the largest-scale evaluation of embedding stability we are aware of.

All three corpora were tokenized, transformed to lower case and cleaned from punctuation. We used both the corpora as-is, as well as independently drawn random subsamples (see also \citet{Hellrich16latech,Antoniak18}) to simulate the arbitrary content selection in most corpora---texts could be removed or replaced with similar ones without changing the overall nature of a corpus, e.g., Wikipedia articles are continuously edited. Subsampling allows us to quantify the effect of this arbitrariness on the stability of embeddings, i.e., how consistently word embeddings are trained on variations of a corpus. Subsampling was performed on the level of the constituent texts of each corpus, e.g., individual news articles. For a corpus with $n$ texts we drew $n$ samples with replacement. Texts could be drawn multiple times, but only one copy was kept, reducing corpora to $1 - \sfrac{1}{e} \approx \sfrac{2}{3}$ of their original size.
\section{Experimental Set-up}
\label{sec:setup}
We compared five algorithm variants: \glove , \sgns , \svdpmi\ without down-sampling, \svdpmi\ with probabilistic down-sampling, and \svdwpmi . While we could use \sgns\footnote{
\myurl{github.com/tmikolov/word2vec}} and \glove\footnote{
\myurl{github.com/stanfordnlp/GloVe}} implementations directly, we had to modify \svdpmi\footnote{
\myurl{github.com/hellrich/hyperwords} -- See also further experimental code: \myurl{github.com/hellrich/embedding_downsampling_comparison}} to support the weighted sampling used in \svdwpmi . As proposed by \citet{Antoniak18}, we further modified our \svdpmi\ implementation to use random numbers generated with a non-fixed seed for probabilistic down-sampling. \textcolor{red}{A fixed seed would benefit reliability, but also act as a bias during all analyses---seed choice has been shown to cause significant differences in experimental results \citep{Henderson18}.}

Down-sampling strategies for $df$ and $f\!\!f$ can be chosen independently of each other, e.g., using probabilistic down-sampling for $df$ together with weighted down-sampling for $f\!\!f$. However, we decided to use the same down-sampling strategies, e.g., weighting, for both factors, taking into account computational limitations as well as results from pre-tests that revealed little benefit of mixed strategies.\footnote{
The strongest counterexample is a combination of probabilistic down-sampling for $df$ and weighting for $f\!\!f$ which lead to small, yet significant improvements in the MEN ($0.703\pm0.001$) and MTurk ($0.568\pm0.015$) similarity tasks (cf.\ Table \ref{tab:results-short}). However, other accuracy tasks showed no improvements and the stability of this approach ($0.475\pm0.001$) was far closer to \svdpmi\ with fully probabilistic down-sampling than to the perfect stability of \svdwpmi .}

\textcolor{red}{We trained ten models for each algorithm variant and corpus.\footnote{
 Hyperparameters roughly follow \citet{Levy15}. We used symmetric 5 word context windows for all models as well as frequent word down-sampling thresholds of $100$ (\glove) and $10^{-4}$ (others). Default learning rates and numbers of iterations were used for all models. Eigenvalues as well as context vectors were ignored for \svdpmi\ embeddings. 5 negative samples were used for SGNS. The minimum frequency threshold was 50 for COHA, 100 for NEWS and 750 for WIKI---increased thresholds were necessary due to \svdpmi 's memory consumption scaling quadratically with vocabulary size. 
}  In the case of subsampling, each model was trained on one of the independently drawn samples. Stability was evaluated by selecting the 1k most frequent words in each non-bootstrap subsampled} corpus as anchor words and calculating $j@10$ (see Equation \ref{eq:rel:jaccard}).\footnote{
Stability calculation was not performed directly between all 10 models, as this would result in a single value and preclude significance tests. Instead, we generated ten j@10 values by calculating the stability of all subsets formed by leaving out each model once in a jackknife procedure.}

Following \citet{Hellrich16latech,Hellrich16coling}, we did not only investigate stability, but also the accuracy of our models to gauge potential trade-offs. 
We measured the Spearman rank correlation between cosine-based word similarity judgments and human ones with four psycholinguistic test sets, i.e., the two crowdsourced test sets MEN \citep{Bruni12} and MTurk \citep{Radinsky11}, the especially strict SimLex-999 \citep{Hill14} and the widely used WordSim-353 (WS-353; \citet{Finkelstein02}). We also measured the percentage of correctly solved analogies (using the multiplicative formula from \newcite{Levy14conll}) with two test sets developed at Google \citep{Mikolov13iclr} and Microsoft Research (MSR; \citet{Mikolov13naacl}). 
\begin{table*}[h]
\footnotesize 
\centering
\begin{tabular}{|c|c|c|c|c|c|c|c|c|c|c|}
\hline
\multirow{2}{*}{Corpus} & \multirow{2}{*}{Algorithm} &   Down-  &  \multicolumn{4}{c|}{Word Similarity} & \multicolumn{2}{c|}{Analogy}  & \multirow{2}{*}{Stability} \\ 
 &  &  sampling & MEN & MTurk & SimLex & WS-353 & Google & MSR &  \\ \hline
 
\multirow{5}{*}{COHA} & \multirow{3}{*}{\svdpmi} & none & 0.697 & \textbf{0.582} & 0.318 & 0.591 & 0.248 & 0.226 & \textbf{1.000}\\
&   & prob. & 0.689 & \textbf{0.571} & 0.333 & 0.577 & 0.224 & 0.257 & 0.324\\
&   & weight & \textbf{0.702} & 0.551 & 0.351 & \textbf{0.594} & \textbf{0.262} & 0.277 & \textbf{1.000}\\
& \multirow{1}{*}{\sgns} & prob. & 0.642 & 0.560 & \textbf{0.394} & 0.551 & 0.248 & \textbf{0.311} & 0.288\\
& \glove & weight & 0.590 & 0.522 & 0.222 & 0.405 & 0.167 & 0.214 & 0.808\\

\hline
\multirow{5}{*}{{\parbox{1cm}{\center COHA Subs.}}} & \multirow{3}{*}{\svdpmi} & none & 0.645 & \textbf{0.537} & 0.267 & \textbf{0.569} & 0.192 & 0.184 & 0.310\\
&   & prob. & 0.632 & \textbf{0.519} & 0.287 & 0.542 & 0.169 & 0.203 & 0.198\\
&   & weight & \textbf{0.651} & \textbf{0.534} & 0.305 & \textbf{0.568} & \textbf{0.206} & 0.235 & 0.329\\
& \multirow{1}{*}{\sgns} & prob. & 0.551 & 0.486 & \textbf{0.363} & 0.479 & 0.192 & \textbf{0.243} & 0.091\\
& \glove & weight & 0.518 & 0.470 & 0.182 & 0.383 & 0.120 & 0.165 & \textbf{0.330}\\
 \hline
 
\multirow{5}{*}{NEWS} & \multirow{3}{*}{\svdpmi} & none & 0.775 & 0.559 & 0.406 & 0.643 & 0.469 & 0.357 & \textbf{1.000}\\
&   & prob. & 0.784 & 0.561 & 0.431 & 0.666 & 0.492 & 0.445 & 0.654\\
&   & weight & \textbf{0.786} & 0.568 & \textbf{0.435} & 0.667 & 0.502 & 0.444 & \textbf{1.000}\\
& \multirow{1}{*}{\sgns} & prob. & 0.739 & \textbf{0.675} & 0.430 & \textbf{0.672} & \textbf{0.643} & \textbf{0.553} & 0.652\\
& \glove & weight & 0.698 & 0.576 & 0.309 & 0.536 & 0.548 & 0.444 & 0.679\\

\hline
 \multirow{5}{*}{{\parbox{1cm}{\center NEWS Subs.}}} & \multirow{3}{*}{\svdpmi} & none & 0.771 & 0.558 & 0.401 & 0.623 & 0.445 & 0.335 & 0.584\\
&   & prob. & 0.776 & 0.564 & 0.423 & 0.642 & 0.463 & 0.420 & 0.571\\
&   & weight & \textbf{0.781} & 0.567 & \textbf{0.430} & \textbf{0.649} & 0.476 & 0.421 & \textbf{0.635}\\
& \multirow{1}{*}{\sgns} & prob. & 0.734 & \textbf{0.673} & 0.417 & \textbf{0.647} & \textbf{0.601} & \textbf{0.513} & 0.452\\
& \glove & weight & 0.687 & 0.572 & 0.301 & 0.508 & 0.505 & 0.408 & 0.461\\
 \hline
 
 \multirow{5}{*}{WIKI} 
& \multirow{3}{*}{\svdpmi} & none & 0.731 & 0.510 & 0.353 & 0.715 & 0.432 & 0.246 & \textbf{1.000}\\
&   & prob. & \textbf{0.747} & 0.571 & 0.392 & \textbf{0.718} & 0.482 & 0.311 & 0.714\\
&   & weight & 0.743 & 0.560 & \textbf{0.393} & \textbf{0.717} & 0.482 & 0.305 & \textbf{1.000}\\
& \multirow{1}{*}{\sgns} & prob. & 0.735 & \textbf{0.659} & 0.372 & \textbf{0.717} & \textbf{0.669} & \textbf{0.421} & 0.488\\
& \glove & weight & 0.744 & 0.651 & 0.354 & 0.667 & 0.653 & 0.397 & 0.666\\

\hline
 \multirow{5}{*}{{\parbox{1cm}{\center WIKI Subs.}}} 
& \multirow{3}{*}{\svdpmi} & none & 0.726 & 0.526 & 0.355 & 0.699 & 0.410 & 0.244 & 0.635\\
&   & prob. & \textbf{0.742} & 0.568 & \textbf{0.391} & \textbf{0.706} & 0.448 & 0.304 & 0.604\\
&   & weight & 0.740 & 0.555 & \textbf{0.389} & \textbf{0.704} & 0.451 & 0.300 & \textbf{0.651}\\
& \multirow{1}{*}{\sgns} & prob. & 0.723 & \textbf{0.657} & 0.364 & 0.686 & \textbf{0.629} & \textbf{0.407} & 0.501\\
 & \glove & weight & 0.735 & 0.642 & 0.345 & 0.655 & 0.599 & 0.382 & 0.486\\
 \hline
 
\end{tabular}
\caption{Performance of different algorithms and down-sampling strategies with models trained on corpora with and without subsampling. \textbf{Bold} values are best or not significantly different by independent t-tests (with $p < 0.05$).
}\label{tab:results-short}
\end{table*} 

\section{Experimental Results}
\label{sec:results}


\begin{figure*}[bt]
    \centering
    \includegraphics[width=.95\textwidth]{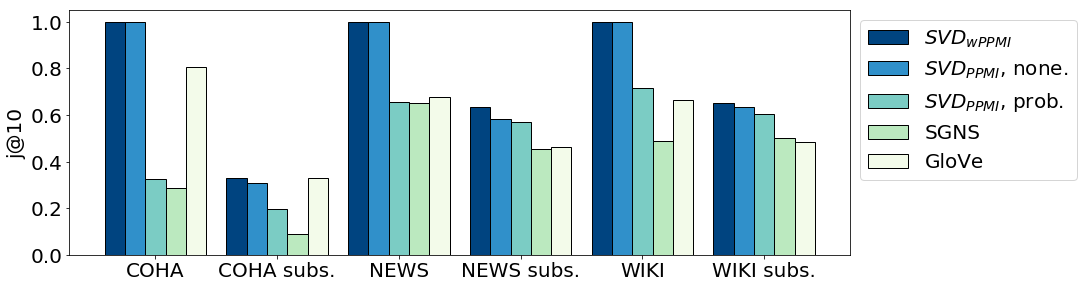}
  
  	\caption{Stability for each combination of algorithm variant and corpus. Measured with j@10 metric (higher is better). Same data as in Table \ref{tab:results-short}, standard deviations too small to \textcolor{red}{display}.\label{fig:rel}}
\end{figure*}

Table \ref{tab:results-short} shows the accuracy and stability for all tested combinations of algorithm and corpus variants. Accuracy differences between test sets are in line with prior observations and general performance on WIKI is roughly in-line with the data reported in \citet{Levy15}.

In general, corpus size does seem to have a positive effect on accuracy. However, for MEN, MTurk and MSR the highest values are achieved with NEWS and not with WIKI. \svdpmi\ variants seem to be less hampered by small training corpora, matching observations by \citet{Sahlgren16}.
Stability is clearly positively influenced by corpus size for all probabilistic algorithm variants except \glove , which, in contrast, benefits from small training corpora. Also, randomly subsampling corpora has a negative effect on both accuracy and stability---this can be explained by the smaller corpus size for accuracy and the differences in training material (as subsampling was performed independently for each model) for stability.

Figure \ref{fig:rel} illustrates the stability of all tested algorithm variants. \svdwpmi\ and \svdpmi\ without down-sampling are the only systems which achieve perfect stability when trained on non-subsampled corpora. \glove\ is the third most reliable algorithm in this scenario, except for the large WIKI corpus. Corpus subsampling decreases the stability of all algorithms, with \svdwpmi\ still performing better than all other alternatives. The only exception is subsampled COHA where the otherwise suboptimal \glove\ narrowly (0.330 instead of 0.329; difference significant with $p<.05$ by two-sided t-test) outperforms \svdwpmi\ . \svdwpmi\ can achieve stability values on subsampled corpora that are competitive with those for \sgns\ and \glove\ trained on \textbf{non}-subsampled corpora. We found standard deviations for stability to be very low, the highest being $0.01$ for \glove\ trained on non-subsampled WIKI, probably due to the overlap in our jackknife procedure.

Finally, we tested\footnote{
All tests were conducted on the averaged accuracy values of the ten individual models per corpus (both subsampled and as-is) and algorithm variant (as listed in Table \ref{tab:results-short}). Using the models directly would have been ill-advised because of their overlapping training data (see \citet[p.\,15]{Demsar06}). Analyses on individual corpora would have resulted in insufficient samples given the pre-conditions of our tests.
} the overall performance of each algorithm variant by first performing a Quade test \citep{Quade79} \textcolor{red}{as a safeguard against type I errors, thus confirming 
the existence of significant differences between algorithms ($p\!=\!1.3 \cdot10^{-7}$)}. 
We then used a pairwise Wilcoxon rank-sum test
 with Holm-\v{S}id\'{a}k correction
 (see \citet{Demsar06}) in order to compare other algorithms with \svdwpmi .\footnote{ This test is a non-parametric alternative to the t-test; corrections prevent false results due to multiple comparisons.} We found it to be not significantly different in accuracy from \sgns\ ($p\!=\!0.101$), but significantly better than \svdpmi\ without down-sampling (corrected $p\!=\!5.4 \cdot 10^{-6}$) 
or probabilistic down-sampling (corrected $p\!=\!0.015$), as well as \glove\ (corrected $p\!=\!0.027$).

Our results show \svdwpmi\ to be both highly reliable and accurate, especially on COHA, which has a size common in both stability studies and corpus linguistic applications. Diverging reports on \svdpmi\ stability---described as perfectly reliable in \citet{Hellrich17dh}, yet not in \citet{Antoniak18}---can thus be explained by their difference in down-sampling options, i.e., no down-sampling or probabilistic down-sampling. \glove 's high stability in other studies \citep{Antoniak18,Wendlandt18} seems to be counterbalanced by its low accuracy and also appears to be limited to training on small corpora.
\section{Discussion}
\label{sec:discussion}
We investigated the effect of down-sampling strategies on word embedding stability by comparing five algorithm variants on three corpora, two of which were larger than those typically used in stability studies. We proposed a simple modification to the down-sampling strategy used for the \svdpmi\ algorithm, \svdwpmi , which uses weighting, to achieve an otherwise unmatched combination of accuracy and stability. We also gathered evidence that \glove\  lacks accuracy and is only  stable when trained on small corpora.

We thus recommend using \svdwpmi , especially for studies targeting (qualitative) interpretations of semantic spaces (e.g., \citet{Kim14}). Overall, SGNS provided no benefit in accuracy over \svdwpmi\ and the latter seemed especially well-suited for small training corpora. The only downside of \svdwpmi\ we are aware of is its relatively large memory consumption during training shared by all \svdpmi\ variants.

Further research could investigate the performance of \svdwpmi\ with other sets of hyperparameters or scrutinize the effect of down-sampling strategies on the ill-understood geometry of embedding spaces \citep{Mimno17}. It would also be interesting to investigate the effect of down-sampling and stability on downstream tasks in a follow-up to \citet{Wendlandt18}.

Finally, the increasingly popular contextualized embedding algorithms, e.g., BERT \citep{Devlin18} or ELMo \citep{Peters18}, are also probabilistic in nature and should thus be affected by stability problems. A direct transfer of our type specific evaluation strategy is impossible. However, an indirect one could be achieved by averaging token-specific contextualized embeddings to generate type representations.
 

\section*{Acknowledgments} \vspace*{-5pt}
This work was supported by the German Federal Ministry of Education and Research (BMBF) within the \textit{SMITH} project (grant 01ZZ1803G),  Deutsche Forschungsgemeinschaft (DFG) within the \textit{STAKI\textsuperscript{2}B\textsuperscript{2}} project (grant \mbox{HA 2097/8-1)}, the  SFB \textit{AquaDiva} (CRC 1076) and the Graduate School \textit{The Romantic Model} (GRK 2041/1). \vspace*{20pt}

\bibliographystyle{acl_natbib}
\bibliography{literature}

\begin{thebibliography}{47}
\expandafter\ifx\csname natexlab\endcsname\relax\def\natexlab#1{#1}\fi

\bibitem[{Antoniak and Mimno(2018)}]{Antoniak18}
Maria Antoniak and David Mimno. 2018.
\newblock Evaluating the stability of embedding-based word similarities.
\newblock \emph{Transactions of the Association for Computational Linguistics},
  6:107--120.

\bibitem[{Berry(1992)}]{Berry92}
M.~W. Berry. 1992.
\newblock Large-scale sparse singular value computations.
\newblock \emph{The International Journal of Supercomputer Applications},
  6(1):13--49.

\bibitem[{Bruni et~al.(2012)Bruni, Boleda, Baroni, and Tran}]{Bruni12}
Elia Bruni, Gemma Boleda, Marco Baroni, and Nam~Khanh Tran. 2012.
\newblock Distributional semantics in technicolor.
\newblock In \emph{ACL 2012 --- Proceedings of the 50th Annual Meeting of the
  Association for Computational Linguistics: Long Papers. Jeju Island, Republic
  of Korea, July 8--14, 2012}, pages 136--145.

\bibitem[{Bullinaria and Levy(2007)}]{Bullinaria07}
John~A. Bullinaria and Joseph~P. Levy. 2007.
\newblock Extracting semantic representations from word co-occurrence
  statistics: a computational study.
\newblock \emph{Behavior Research Methods}, 39(3):510--526.

\bibitem[{Chugh et~al.(2018)Chugh, Whigham, and Dick}]{Chugh18}
Mansi Chugh, Peter~A. Whigham, and Grant Dick. 2018.
\newblock Stability of word embeddings using word2vec.
\newblock In \emph{Advances in Artificial Intelligence. AI 2018 --- Proceedings
  of the 31st Australasian Joint Conference on Artificial Intelligence.
  Wellington, New Zealand. December 11-14, 2018}, pages 812--818.

\bibitem[{Church and Hanks(1990)}]{Church90}
Kenneth~Ward Church and Patrick Hanks. 1990.
\newblock Word association norms, mutual information, and lexicography.
\newblock \emph{Computational Linguistics}, 16(1):22--29.

\bibitem[{Coiera et~al.(2018)Coiera, Ammenwerth, Georgiou, and
  Magrabi}]{Coiera18}
Enrico Coiera, Elske Ammenwerth, Andrew Georgiou, and Farah Magrabi. 2018.
\newblock Does health informatics have a replication crisis?
\newblock \emph{Journal of the American Medical Informatics Association},
  25(8):963--968.

\bibitem[{Davies(2012)}]{Davies12}
Mark Davies. 2012.
\newblock Expanding horizons in historical linguistics with the 400-million
  word {Corpus of Historical American English}.
\newblock \emph{Corpora}, 7:121--157.

\bibitem[{Dem{\v{s}}ar(2006)}]{Demsar06}
Janez Dem{\v{s}}ar. 2006.
\newblock Statistical comparisons of classifiers over multiple data sets.
\newblock \emph{Journal of Machine Learning Research}, 7:1--30.

\bibitem[{Devlin et~al.(2018)Devlin, Chang, Lee, and Toutanova}]{Devlin18}
Jacob Devlin, Ming{-}Wei Chang, Kenton Lee, and Kristina Toutanova. 2018.
\newblock {BERT:} pre-training of deep bidirectional transformers for language
  understanding.
\newblock \url{http://arxiv.org/abs/1810.04805}.

\bibitem[{Fano(1961)}]{Fano61}
Robert~M. Fano. 1961.
\newblock \emph{Transmission of Information. A Statistical Theory of
  Communications}.
\newblock MIT Press.
\newblock 3rd printing, 1966.

\bibitem[{Finkelstein et~al.(2002)Finkelstein, Gabrilovich, Matias, Rivlin,
  Solan, Wolfman, and Ruppin}]{Finkelstein02}
Lev Finkelstein, Evgeniy Gabrilovich, Yossi Matias, Ehud Rivlin, Zach Solan,
  Gadi Wolfman, and Eytan Ruppin. 2002.
\newblock Placing search in context: the concept revisited.
\newblock \emph{ACM Transactions on Information Systems}, 20(1):116--131.

\bibitem[{Halko et~al.(2011)Halko, Martinsson, and Tropp}]{Halko11}
Nathan Halko, Per-Gunnar Martinsson, and Joel~A. Tropp. 2011.
\newblock Finding structure with randomness: probabilistic algorithms for
  constructing approximate matrix decompositions.
\newblock \emph{SIAM Review}, 53(2):217--288.

\bibitem[{Harris(1954)}]{Harris54}
Zellig~S. Harris. 1954.
\newblock Distributional structure.
\newblock \emph{Word}, 10(2-3):146--162.

\bibitem[{Hellrich et~al.(2018)Hellrich, Buechel, and Hahn}]{Hellrich18coling}
Johannes Hellrich, Sven Buechel, and Udo Hahn. 2018.
\newblock \textsc{JeSemE}: a website for exploring diachronic changes in word
  meaning and emotion.
\newblock In \emph{COLING 2018 --- Proceedings of the 27th International
  Conference on Computational Linguistics: System Demonstrations. Santa Fe, NM,
  USA, August 20--26, 2018}, pages 10--14.

\bibitem[{Hellrich and Hahn(2016{\natexlab{a}})}]{Hellrich16latech}
Johannes Hellrich and Udo Hahn. 2016{\natexlab{a}}.
\newblock An assessment of experimental protocols for tracing changes in word
  semantics relative to accuracy and reliability.
\newblock In \emph{LaTeCH 2016 --- Proceedings of the 10th SIGHUM Workshop on
  Language Technology for Cultural Heritage, Social Sciences, and Humanities @
  ACL 2016, Berlin, Germany, August 11, 2016}, pages 111--117.

\bibitem[{Hellrich and Hahn(2016{\natexlab{b}})}]{Hellrich16coling}
Johannes Hellrich and Udo Hahn. 2016{\natexlab{b}}.
\newblock Bad company: neighborhoods in neural embedding spaces considered
  harmful.
\newblock In \emph{COLING 2016 --- Proceedings of the 26th International
  Conference on Computational Linguistics: Technical Papers. Osaka, Japan,
  December 11--16, 2016}, pages 2785--2796.

\bibitem[{Hellrich and Hahn(2017)}]{Hellrich17dh}
Johannes Hellrich and Udo Hahn. 2017.
\newblock Don't get fooled by word embeddings: better watch their neighborhood.
\newblock In \emph{Digital Humanities 2017 --- Conference Abstracts of the 2017
  Conference of the Alliance of Digital Humanities Organizations (ADHO).
  Montr\'{e}al, Quebec, Canada, August 8--11, 2017}, pages 250--252.

\bibitem[{Henderson et~al.(2018)Henderson, Islam, Bachman, Pineau, Precup, and
  Meger}]{Henderson18}
Peter Henderson, Riashat Islam, Philip Bachman, Joelle Pineau, Doina Precup,
  and David Meger. 2018.
\newblock Deep reinforcement learning that matters.
\newblock In \emph{AAAI-IAAI-EAAI '18 --- Proceedings of the 32nd AAAI
  Conference on Artificial Intelligence \& 30th Conference on Innovative
  Applications of Artificial Intelligence \& 8th Symposium on Educational
  Advances in Artificial Intelligence. New Orleans, Louisiana, USA, February
  2-7, 2018}, pages 3207--3214.

\bibitem[{Hill et~al.(2014)Hill, Reichart, and Korhonen}]{Hill14}
Felix Hill, Roi Reichart, and Anna Korhonen. 2014.
\newblock {SimLex-999}: Evaluating semantic models with (genuine) similarity
  estimation.
\newblock \emph{Computational Linguistics}, 41(4):665--695.

\bibitem[{Ivie and Thain(2018)}]{Ivie18}
Peter Ivie and Douglas Thain. 2018.
\newblock Reproducibility in scientific computing.
\newblock \emph{ACM Computing Surveys}, 51(3):63:1--63:36.

\bibitem[{Jaccard(1912)}]{Jaccard12}
Paul Jaccard. 1912.
\newblock The distribution of the flora in the alpine zone.
\newblock \emph{New Phytologist}, XI(2):37--50.
\newblock [Translation of 1901 article].

\bibitem[{Kim et~al.(2014)Kim, Chiu, Hanaki, Hegde, and Petrov}]{Kim14}
Yoon Kim, Yi-I Chiu, Kentaro Hanaki, Darshan Hegde, and Slav Petrov. 2014.
\newblock Temporal analysis of language through neural language models.
\newblock In \emph{Proceedings of the Workshop on Language Technologies and
  Computational Social Science @ ACL 2014. Baltimore, Maryland, USA, June 26,
  2014}, pages 61--65.

\bibitem[{Kulkarni et~al.(2015)Kulkarni, Al-Rfou, Perozzi, and
  Skiena}]{Kulkarni15}
Vivek Kulkarni, Rami Al-Rfou, Bryan Perozzi, and Steven Skiena. 2015.
\newblock Statistically significant detection of linguistic change.
\newblock In \emph{WWW 2015 --- Proceedings of the 24th International
  Conference on World Wide Web: Technical Papers. Florence, Italy, May 18--22,
  2015}, pages 625--635.

\bibitem[{Levy and Goldberg(2014{\natexlab{a}})}]{Levy14acl}
Omer Levy and Yoav Goldberg. 2014{\natexlab{a}}.
\newblock Dependency-based word embeddings.
\newblock In \emph{ACL 2014 --- Proceedings of the 52nd Annual Meeting of the
  Association for Computational Linguistics: Short Papers. Baltimore, Maryland,
  USA, June 22--27, 2014}, pages 302--308.

\bibitem[{Levy and Goldberg(2014{\natexlab{b}})}]{Levy14conll}
Omer Levy and Yoav Goldberg. 2014{\natexlab{b}}.
\newblock Linguistic regularities in sparse and explicit word representations.
\newblock In \emph{CoNLL 2014 --- Proceedings of the 18th Conference on
  Computational Natural Language Learning @ ACL 2014. Baltimore, Maryland, USA,
  June 26-27, 2014}, pages 171--180.

\bibitem[{Levy and Goldberg(2014{\natexlab{c}})}]{Levy14nips}
Omer Levy and Yoav Goldberg. 2014{\natexlab{c}}.
\newblock Neural word embedding as implicit matrix factorization.
\newblock In \emph{Advances in Neural Information Processing Systems 27 ---
  NIPS 2014. Proceedings of the 28th Annual Conference on Neural Information
  Processing Systems 2014. Montr\'{e}al, Qu\'{e}bec, Canada, December 8-13,
  2014}, pages 2177--2185.

\bibitem[{Levy et~al.(2015)Levy, Goldberg, and Dagan}]{Levy15}
Omer Levy, Yoav Goldberg, and Ido Dagan. 2015.
\newblock Improving distributional similarity with lessons learned from word
  embeddings.
\newblock \emph{Transactions of the Association for Computational Linguistics},
  3:211--225.

\bibitem[{Michel et~al.(2011)Michel, Shen, Aiden, Veres, Gray, Google
  Books~Team, Pickett, Hoiberg, Clancy, Norvig, Orwant, Pinker, Nowak, and
  Aiden}]{Michel11}
Jean-Baptiste Michel, Yuan~Kui Shen, Aviva~Presser Aiden, Adrian Veres,
  Matthew~K. Gray, The Google Books~Team, Joseph~P. Pickett, Dale Hoiberg, Dan
  Clancy, Peter Norvig, Jon Orwant, Steven Pinker, Martin~A. Nowak, and
  Erez~Lieberman Aiden. 2011.
\newblock Quantitative analysis of culture using millions of digitized books.
\newblock \emph{Science}, 331(6014):176--182.

\bibitem[{Mikolov et~al.(2013{\natexlab{a}})Mikolov, Chen, Corrado, and
  Dean}]{Mikolov13iclr}
Tomas Mikolov, Kai Chen, Gregory~S. Corrado, and Jeffrey Dean.
  2013{\natexlab{a}}.
\newblock Efficient estimation of word representations in vector space.
\newblock In \emph{ICLR 2013 --- Workshop Proceedings of the International
  Conference on Learning Representations. Scottsdale, Arizona, USA, May 2--4,
  2013}.
\newblock \url{https://arxiv.org/abs/1301.3781}.

\bibitem[{Mikolov et~al.(2013{\natexlab{b}})Mikolov, Yih, and
  Zweig}]{Mikolov13naacl}
Tomas Mikolov, Wen-tau Yih, and Geoffrey Zweig. 2013{\natexlab{b}}.
\newblock Linguistic regularities in continuous space word representations.
\newblock In \emph{NAACL-HLT 2013 --- Proceedings of the 2013 Conference of the
  North American Chapter of the Association for Computational Linguistics:
  Human Language Technologies. Atlanta, GA, USA, 9--14 June 2013}, pages
  746--751.

\bibitem[{Mikolov et~al.(2013{\natexlab{c}})Mikolov, Sutskever, Chen, Corrado,
  and Dean}]{Mikolov13nips}
Tom\'{a}\v{s} Mikolov, Ilya Sutskever, Kai Chen, Gregory~S. Corrado, and
  Jeffrey Dean. 2013{\natexlab{c}}.
\newblock Distributed representations of words and phrases and their
  compositionality.
\newblock In \emph{Advances in Neural Information Processing Systems 26 ---
  NIPS 2013. Proceedings of the 27th Annual Conference on Neural Information
  Processing Systems. Lake Tahoe, Nevada, USA, December 5-10, 2013}, pages
  3111--3119.

\bibitem[{Mimno and Thompson(2017)}]{Mimno17}
David Mimno and Laure Thompson. 2017.
\newblock The strange geometry of skip-gram with negative sampling.
\newblock In \emph{EMNLP 2017 --- Proceedings of the 2017 Conference on
  Empirical Methods in Natural Language Processing. Copenhagen, Denmark,
  September 7--11, 2017}, pages 2863--2868.

\bibitem[{Niwa and Nitta(1994)}]{Niwa94}
Yoshiki Niwa and Yoshihiko Nitta. 1994.
\newblock Co-occurrence vectors from corpora vs. distance vectors from
  dictionaries.
\newblock In \emph{COLING 1994 --- Proceedings of the 15th Conference on
  Computational Linguistics: Volume 1. Kyoto, Japan, August 5--9, 1994}, pages
  304--309.

\bibitem[{Padr\'o et~al.(2014)Padr\'o, Idiart, Villavicencio, and
  Ramisch}]{Padro14}
Muntsa Padr\'o, Marco Idiart, Aline Villavicencio, and Carlos Ramisch. 2014.
\newblock Comparing similarity measures for distributional thesauri.
\newblock In \emph{LREC 2014 --- Proceedings of the 9th International
  Conference on Language Resources and Evaluation. Reykjavik, Iceland, May
  26-31, 2014}, pages 2694--2971.

\bibitem[{Pennington et~al.(2014)Pennington, Socher, and
  Manning}]{Pennington14}
Jeffrey Pennington, Richard Socher, and Christopher~D. Manning. 2014.
\newblock {GloVe}: Global vectors for word representation.
\newblock In \emph{EMNLP 2014 --- Proceedings of the 2014 Conference on
  Empirical Methods in Natural Language Processing. Doha, Qatar, October
  25--29, 2014}, pages 1532--1543.

\bibitem[{Peters et~al.(2018)Peters, Neumann, Iyyer, Gardner, Clark, Lee, and
  Zettlemoyer}]{Peters18}
Matthew~E. Peters, Mark Neumann, Mohit Iyyer, Matt Gardner, Christopher~T.
  Clark, Kenton Lee, and Luke~S. Zettlemoyer. 2018.
\newblock Deep contextualized word representations.
\newblock In \emph{NAACL-HLT 2018 --- Proceedings of the 2018 Conference of the
  North American Chapter of the Association for Computational Linguistics:
  Human Language Technologies. New Orleans, Louisiana, USA, June 1-6, 2018},
  volume 1: Long Papers, pages 2227--2237.

\bibitem[{Pierrejean and Tanguy(2018)}]{Pierrejean18taln}
B\'{e}n\'{e}dicte Pierrejean and Ludovic Tanguy. 2018.
\newblock \'{E}tude de la reproductibilit\'{e} des word embeddings:
  rep\'{e}rage des zones stables et instables dans le lexique.
\newblock In \emph{TALN 2018 --- Actes de la 25\`{e}me conf\'{e}rence sur le
  Traitement Automatique des Langues Naturelles. Rennes, France, 14-18 Mai,
  2018.}, volume 1: Articles longs, articles courts de TALN, pages 33--46.

\bibitem[{Quade(1979)}]{Quade79}
Dana Quade. 1979.
\newblock Using weighted rankings in the analysis of complete blocks with
  additive block effects.
\newblock \emph{Journal of the American Statistical Association},
  74(367):680--683.

\bibitem[{Radinsky et~al.(2011)Radinsky, Agichtein, Gabrilovich, and
  Markovitch}]{Radinsky11}
Kira Radinsky, Eugene Agichtein, Evgeniy Gabrilovich, and Shaul Markovitch.
  2011.
\newblock A word at a time: computing word relatedness using temporal semantic
  analysis.
\newblock In \emph{WWW 2011 --- Proceedings of the 20th International
  Conference on World Wide Web. Hyderabad, India, March 28 - April 1, 2011},
  pages 337--346.

\bibitem[{Reimers and Gurevych(2017)}]{Reimers17}
Nils Reimers and Iryna Gurevych. 2017.
\newblock Reporting score distributions makes a difference: performance study
  of {LSTM-networks} for sequence tagging.
\newblock In \emph{EMNLP 2017 --- Proceedings of the 2017 Conference on
  Empirical Methods in Natural Language Processing. Copenhagen, Denmark,
  September 9-11, 2017}, pages 338--348.

\bibitem[{Rubenstein and Goodenough(1965)}]{Rubenstein65}
Herbert Rubenstein and John~B. Goodenough. 1965.
\newblock Contextual correlates of synonymy.
\newblock \emph{Communications of the ACM}, 8(10):627--633.

\bibitem[{Saad(2003)}]{Saad03}
Yousef Saad. 2003.
\newblock \emph{Iterative Methods for Sparse Linear Systems}, 2nd edition.
\newblock Society for Industrial and Applied Mathematics, Philadelphia/PA.

\bibitem[{Sahlgren and Lenci(2016)}]{Sahlgren16}
Magnus Sahlgren and Alessandro Lenci. 2016.
\newblock The effects of data size and frequency range on distributional
  semantic models.
\newblock In \emph{EMNLP 2016 --- Proceedings of the 2016 Conference on
  Empirical Methods in Natural Language Processing. Austin, Texas, USA,
  November 1--5, 2016}, pages 975--980.

\bibitem[{Salton and Lesk(1971)}]{Salton71chapter}
Gerald Salton and Michael~E. Lesk. 1971.
\newblock Information analysis and dictionary construction.
\newblock In Gerald Salton, editor, \emph{The \textsc{Smart} Retrieval System:
  Experiments in Automatic Document Processing}, chapter~6, pages 115--142.
  Prentice-Hall, Englewood Cliffs/NJ.

\bibitem[{Weeds et~al.(2004)Weeds, Weir, and McCarthy}]{Weeds04}
Julie Weeds, David Weir, and Diana McCarthy. 2004.
\newblock Characterising measures of lexical distributional similarity.
\newblock In \emph{COLING 2004 --- Proceedings of the 20th International
  Conference on Computational Linguistics. Geneva, Switzerland, Aug 23--27,
  2004}, pages 1015--1021.

\bibitem[{Wendlandt et~al.(2018)Wendlandt, Kummerfeld, and
  Mihalcea}]{Wendlandt18}
Laura Wendlandt, Jonathan~K. Kummerfeld, and Rada Mihalcea. 2018.
\newblock Factors influencing the surprising instability of word embeddings.
\newblock In \emph{NAACL-HLT 2018 --- Proceedings of the 2018 Conference of the
  North American Chapter of the Association for Computational Linguistics:
  Human Language Technologies: Long Papers. New Orleans, LA, USA, June 2--4,
  2018}, pages 2092--2102.

\end{thebibliography}

\end{document}